\documentclass[a4paper,twoside]{article}

\usepackage{epsfig}
\usepackage{calc}
\usepackage{amssymb}
\usepackage{amstext}
\usepackage{amsmath}
\usepackage{amsthm}
\usepackage{multicol}
\usepackage{times}
\usepackage{apalike}
\usepackage{algorithm2e}
\usepackage[bottom]{footmisc}
\usepackage{SCITEPRESS}

\author{\authorname{Rowan Martnishn}
\affiliation{University of Pennsylvania}
}

\title{Pre-Warm: Initializing Convolutional Filters from First-Batch Patch Dictionaries}


\abstract{Random initialization of convolutional filters does not use the training images.
Previous work has shown that image patches can be copied into the first layer, and that
$k$-means or principal components of patches can serve as filters. This paper compares
four initializations of the first layer of a small convolutional network, with every other
factor held fixed: He initialization, random mean-centered patches, principal components
of those patches, and $k$-means centroids. Pre-Warm, our proposed methodology, is the rule-based use of both dictionaries:
the patch count follows the filter count and a foreground density, both dictionaries
are built from a single minibatch, and whichever of principal components or $k$-means
better reconstructs those patches is written into the first half of the filter bank,
rather than chosen by a validation search. The remaining filters stay random. On five datasets, principal components
improve CIFAR-10 and CIFAR-100 relative to He initialization, and $k$-means improves SVHN and MNIST, and is the stronger of the two on Fashion-MNIST; copying raw patches does not reproduce those color-set gains. Use principal
components on photographic patches and $k$-means on stroke-like patches; the
first-batch reconstruction check recovers that split.}

\begin{document}

\onecolumn \maketitle \normalsize \setcounter{footnote}{0} \vfill

\section{\uppercase{Introduction}}
\label{sec:introduction}

\noindent
Before a convolutional network can be trained, its filters have to be set to some initial
value. The usual choice is a scaled random draw that preserves activation variance
\cite{glorot2010understanding,he2015delving}. That construction is the same for every
dataset: the training images are not used until the first gradient step.

First-layer filters, however, tend to become local edge and texture operators. The same
structure can be read off the training images themselves by clustering or factorizing small
patches \cite{coates2011analysis,chan2015pcanet}. Lehmann and Ebner
\cite{lehmann2021patches} showed, at this conference, that copying such patches into
convolutional filters still allows training and can be more robust to a poor learning rate.
The open question is not whether a patch fill trains, but whether a dictionary built from
those patches is any better than copying the patches, and whether that answer depends on
the image domain.

This paper studies that comparison on a constant, foundational architecture. Four fills of the first layer
are trained under a matched protocol: He initialization; random mean-centered patches;
principal components of those patches; and MiniBatch $k$-means centroids. Pre-Warm is
the rule-based use of the last two: the patch count follows the filter count and a
foreground density, and a first-batch reconstruction check chooses principal components
or $k$-means, rather than a validation search. The last three methods share the same
first-batch sample, the same spatial weighting, and the same assignment to the first
half of the filter bank. The remaining
filters, and all later layers, stay at He initialization. The questions are:

\begin{enumerate}
\item Does a first-batch patch dictionary beat He initialization on a small convolutional
network, under a matched training budget?
\item Is the useful dictionary the same on photographs, house numbers, and grayscale
objects?
\item Is copying raw patches enough, or does our proposed, rule-based dictionary estimator matter?
\end{enumerate}

The result is a domain split, not a uniform gain. Principal components are the stronger
dictionary on CIFAR-10 and CIFAR-100 photographs. $k$-means is the stronger dictionary
on SVHN, MNIST, and Fashion. Unstructured patch copies stay close to He initialization on the color sets.
Fashion-MNIST is not significant. The claim is therefore limited: a first-batch patch dictionary can
help the first layer of a small convolutional network, and the useful dictionary is not the
same on every dataset.

The rest of the paper is organized as follows. Section~\ref{sec:related} reviews random
and patch-based initialization. Section~\ref{sec:method} describes the shared pipeline and the two rules.
Section~\ref{sec:experiments} reports the five-dataset comparison. Section~\ref{sec:conclusion}
concludes.

\section{\uppercase{Related Work}}
\label{sec:related}

\noindent
Glorot and Bengio \cite{glorot2010understanding} and He et al.\ \cite{he2015delving} set
the scale of random weights from fan-in (and, for Xavier, fan-out). Both methods are
distribution-free. Kr\"ahenb\"uhl et al.\ \cite{krahenbuhl2016data} instead adjust random
weights from forward-pass statistics of the training set. That is data-dependent, but it does
not put image content into the filters.

Using patches as convolutional features is older than these initializers. Coates et al.\
\cite{coates2011analysis} showed that $k$-means on whitened patches yields competitive
single-layer features. PCANet uses stacked principal-component filters and histogram
features, without backpropagating through the filters \cite{chan2015pcanet}.
Lehmann and Ebner \cite{lehmann2021patches} copy training patches into convolutional
layers of a network that is then trained as usual. Rabinovich et al.\
\cite{rabinovich2025init} write $k$-means centroids or principal components into the first
layer and treat the patch count and the dictionary class as validation choices.

The present comparison holds the architecture, optimizer, and patch sample fixed, and
changes only how the first half of conv1 is filled. Random patches test whether any
training pixels in the filters are enough. Principal components and $k$-means test two
standard dictionaries on that same sample. The patch count is set from the filter count and
a foreground density rather than a grid search. A first-batch reconstruction check is the deployment
rule for choosing the dictionary class; both dictionaries are still reported on every dataset.

\section{\uppercase{Method}}
\label{sec:method}

\noindent
Let the first convolutional layer have $F$ output channels, $C_{\mathrm{in}}$ input
channels, and kernel size $K$. All experiments use $F=32$ and $K=3$. From the first
training minibatch, $n$ windows of size $K\times K$ are sampled from each image,
flattened, and mean-centered,
\begin{equation}
\tilde{p} \leftarrow p - \bar{p}.
\label{eq:center}
\end{equation}
Mean-centering removes the DC component so that the dictionaries encode spatial structure
(and color, when $C_{\mathrm{in}}=3$) rather than luminance.

Three dictionaries are formed from the same matrix of centered patches $P$. Random
selects $F/2$ rows of $P$ without replacement. PCA takes the leading $F/2$ eigenvectors of
the patch covariance. MiniBatch $k$-means uses $F/2$ centroids. Pre-Warm is the
rule that builds both dictionaries and keeps the one with the lower first-batch
reconstruction error. Each dictionary is then finished in the same way. Element $(i,j)$ of every $K\times K$ spatial
slice is multiplied by the inverse-Manhattan weight
\begin{equation}
w_{ij} = \bigl(1 + |i-\lfloor K/2\rfloor| + |j-\lfloor K/2\rfloor|\bigr)^{-1},
\label{eq:weight}
\end{equation}
which down-weights corners relative to the receptive-field center. The weighted matrix is
rescaled to a target standard deviation $\sigma$,
\begin{equation}
C \leftarrow \frac{\sigma\, C}{\mathrm{std}(C)+\varepsilon},
\label{eq:scale}
\end{equation}
reshaped, and copied into the first $F/2$ output channels of conv1. The remaining $F/2$
filters, and all other layers, keep He initialization \cite{he2015delving}. Batch
normalization follows conv1, so any initialization signal has to survive a later affine
normalization.

He initialization for a ReLU convolution uses standard deviation
$\sqrt{2/\mathrm{fan_{in}}}$. For $3\times3\times3$ color filters that value is about $0.272$.
The color experiments use $\sigma=0.25$. The grayscale experiments use $\sigma=0.20$.
Writing only half of the bank is a heuristic: it injects a data-derived subspace without
replacing the entire random basis \cite{lehmann2021patches,rabinovich2025init}. The
shared fill is:

\begin{enumerate}
\item Sample $n$ patches of size $K\times K$ from each image in the first minibatch.
\item Mean-center each patch as in \eqref{eq:center}.
\item Form a dictionary of size $F/2$: random rows of $P$, principal components, or
MiniBatch $k$-means centroids.
\item Apply \eqref{eq:weight} and rescale with \eqref{eq:scale}.
\item Write the dictionary into the first $F/2$ filters. Leave the rest at He initialization.
\end{enumerate}

\subsection{Patch Count}

Patch size equals the kernel size so that dictionary atoms map onto filters without
cropping. The codebook size is $k=\lfloor F/2\rfloor=16$. Let $d\in(0,1]$ be the fraction of
uniformly drawn patches that are treated as foreground.

$d$ is the logistic of two training-set moments: mean BT.601 luminance
$\mu$ and the mean Euclidean norm $\rho$ of mean-centered $3\times 3$
patches,
\begin{equation}
\begin{gathered}
d=\phi\bigl(84.88-173.53\,\mu-280.99\,\rho+575.02\,\mu\rho\bigr),\\
\phi(t)=(1+e^{-t})^{-1}.
\end{gathered}
\label{eq:density}
\end{equation}

The sampling rate is
\begin{equation}
n=\Bigl\lfloor\frac{F}{4d}\Bigr\rfloor,
\label{eq:occupancy}
\end{equation}
which keeps a roughly constant number of foreground patches per image as $d$ changes.
Background patches, after \eqref{eq:center}, sit near the origin and pull $k$-means toward
zero if they dominate the sample. Based on the occupancy criterion in \eqref{eq:occupancy}, the patch counts were
fixed before training at $n=58$ for MNIST, $n=22$ for Fashion-MNIST, $n=28$
for CIFAR-10, $n=16$ for CIFAR-100, and $n=15$ for SVHN.

\subsection{Choosing the Dictionary}

Principal components and $k$-means reconstruct the same centered patches in different
ways: a $k$-dimensional subspace versus $k$ Voronoi centroids. Both reconstructions can
be scored on the first batch, with no labels. The deployment rule is to form both
dictionaries on $P$ and keep the one with the lower reconstruction residual. Ties default to
principal components. Random patches are not a candidate under this rule; they are only a
control for ``any data in conv1.''

The rule is a convenience for a single deployed initializer. Table~\ref{tab:main} still
reports both dictionaries on every dataset, so the accuracy split can be read independently
of the rule.

\subsection{Summary of the Rules}

\noindent
The deployed initializer is those two rules, not a search. Random patches are only a
control. Ties go to principal components:
\begin{equation}
n=\Bigl\lfloor\frac{F}{4d}\Bigr\rfloor,
\quad
C^{\star}
=
\begin{cases}
C_{\mathrm{PCA}} & \text{if } r_{\mathrm{PCA}}\le r_{\mathrm{VQ}},\\
C_{\mathrm{VQ}} & \text{otherwise.}
\end{cases}
\label{eq:prewarm}
\end{equation}
$C_{\mathrm{PCA}}$ and $C_{\mathrm{VQ}}$ are the principal-component and $k$-means
dictionaries of size $F/2$; $r$ is first-batch reconstruction error.

\begin{small}
\begin{verbatim}
n = floor(F / (4*d))    # patch count
P = mean-centered KxK first-batch patches
C_pca, C_km = PCA(P), kmeans(P) # size F/2
if recon(P, C_pca) <= recon(P, C_km):
    write C_pca into conv1[:F/2]
else:
    write C_km  into conv1[:F/2]
# conv1[F/2:] stays He
\end{verbatim}
\end{small}

\section{\uppercase{Experiments}}
\label{sec:experiments}

\noindent
All runs use one architecture, FlexCNN: a $3\times3$ convolution from $C_{\mathrm{in}}$ to
$32$ channels with padding $1$, batch-norm, ReLU, and $2\times2$ max-pooling; a second
block of the same form to $64$ channels; then fully connected layers of width $256$ and
$N_{\mathrm{cls}}$. The flattened width is $64\times7\times7$ on $28\times28$ inputs and
$64\times8\times8$ on $32\times32$ inputs.

The datasets are MNIST \cite{lecun1998gradient}, Fashion-MNIST \cite{xiao2017fashion},
CIFAR-10 and CIFAR-100 \cite{krizhevsky2009learning}, and cropped SVHN
\cite{netzer2011reading}. Pixels are scaled to $[0,1]$ and standardized with the usual
per-channel mean and standard deviation. Each run trains for $5000$ Adam steps, batch
size $256$, with cosine annealing. Table~\ref{tab:setup} lists the patch count, amplitude,
and learning-rate range used on each dataset. The eight seeds
$\{42,7,13,99,2025,1,8,21\}$ control He draws, shuffling, $k$-means, PCA, and patch
sampling, so the four methods are paired. Test accuracy is recorded at the final step on the
official split. Significance is a one-sided paired $t$-test of each method versus He
initialization over $n=8$ seeds, not over test images.

\begin{table}[t]
\caption{Training and sampling settings. $n$ is the number of $3\times3$ patches drawn from
each image in the first batch. Configuration values follow \eqref{eq:occupancy}.}\label{tab:setup}
\centering
\small
\begin{tabular}{|l|c|c|c|}
\hline
Dataset & $n$ & $\sigma$ & Learning rate \\
\hline
MNIST & 58 & 0.20 & $3\mathrm{e}{-3}\to 3\mathrm{e}{-5}$ \\
Fashion-MNIST & 22 & 0.20 & $3\mathrm{e}{-3}\to 3\mathrm{e}{-5}$ \\
CIFAR-10 & 28 & 0.25 & $3\mathrm{e}{-3}\to 3\mathrm{e}{-5}$ \\
CIFAR-100 & 16 & 0.25 & $2\mathrm{e}{-3}\to 3\mathrm{e}{-5}$ \\
SVHN & 15 & 0.25 & $5\mathrm{e}{-4}\to 3\mathrm{e}{-6}$ \\
\hline
\end{tabular}
\end{table}

\subsection{Comparison to He Initialization}

Table~\ref{tab:main} reports mean test accuracy and paired differences versus He
initialization. At $\alpha=0.05$, PCA exceeds He initialization on CIFAR-10
($\Delta=+0.98$~pp, $p=0.0003$, $8/8$ seeds) and CIFAR-100 ($\Delta=+0.76$~pp,
$p=0.0016$, $8/8$). $k$-means exceeds He initialization on SVHN ($\Delta=+0.56$~pp,
$p=0.0001$, $8/8$), CIFAR-100 ($\Delta=+0.49$~pp, $p=0.035$, $7/8$), and MNIST
($\Delta=+0.05$~pp, $p=0.009$, $7/8$). The MNIST increment sits at a $99.3\%$ ceiling
and is not a practical gain. Fashion-MNIST is not significant for any method ($k$-means
$p=0.096$). CIFAR-10 $k$-means is $p=0.067$. Those two tests sit just above
$\alpha=0.05$; both mean differences are positive, so they agree in sign with the
significant $k$-means results, but with $n=8$ seeds they are not strong enough to
claim an effect.

\begin{table}[t]
\caption{Mean test accuracy (\%, $n=8$ seeds). $\Delta$ is the difference versus He
initialization, in percentage points.}\label{tab:main}
\centering
\small
\setlength{\tabcolsep}{3pt}
\begin{tabular}{|l|c|c|c|c|}
\hline
Dataset & He & Rand.\ $\Delta$ & PCA $\Delta$ & $k$-m.\ $\Delta$ \\
\hline
MNIST & $99.28\pm0.03$ & ${+0.04}$ & $+0.02$ & ${\bf +0.05}$ \\
Fashion & $92.69\pm0.19$ & $+0.02$ & $+0.06$ & ${\bf +0.12}$ \\
CIFAR-10 & $74.44\pm0.33$ & $+0.21$ & ${\bf +0.98}$ & $+0.34$ \\
CIFAR-100 & $43.08\pm0.42$ & $+0.08$ & ${\bf +0.76}$ & $+0.49$ \\
SVHN & $88.98\pm0.18$ & $+0.08$ & $+0.11$ & ${\bf +0.56}$ \\
\hline
\end{tabular}
\end{table}

The absolute CIFAR-10 accuracy near $74\%$ shows that the $5000$-step budget is an
initialization comparison, not a fully trained baseline. The point of the protocol is that
the four fills see the same optimizer, the same seeds, and the same number of gradient
steps. 

\subsection{Random Patches}

Random patches never improve the color datasets ($p\ge 0.16$). On MNIST they match the
same ceiling bump as $k$-means ($p=0.027$). That control matters for reading
Table~\ref{tab:main}. If any training pixels in conv1 were enough, the random fill would
move CIFAR and SVHN with PCA and $k$-means. It does not. The dictionary estimator is
what moves those sets, not the mere presence of image content in the filters.

This is the main difference relative to Lehmann and Ebner \cite{lehmann2021patches}, who
showed that patch copies can train and can be more robust to a poor learning rate. Under
the present matched budget, copying patches is not what produces the various
gains. Averaging (PCA or $k$-means) is.

The split also matches the reconstruction rule in Section~\ref{sec:method} on every
dataset in Table~\ref{tab:main}. The subspace dictionary is the one that moves the
photographic sets, CIFAR-10 and CIFAR-100 (PCA is significant on both; $k$-means is
significant only on CIFAR-100, and is weaker there). The centroid dictionary is the one
that moves the stroke-like digit sets: SVHN and MNIST ($k$-means is significant on both;
PCA is not). Fashion-MNIST sits between those regimes and is not significant for either
dictionary. Natural-image patches are well described by a low-dimensional linear model.
Digit patches are more stroke-like, and a small set of centroids can represent that mass
more cheaply than a subspace of the same size.

\begin{figure*}[t]
\centering
\includegraphics[width=\textwidth]{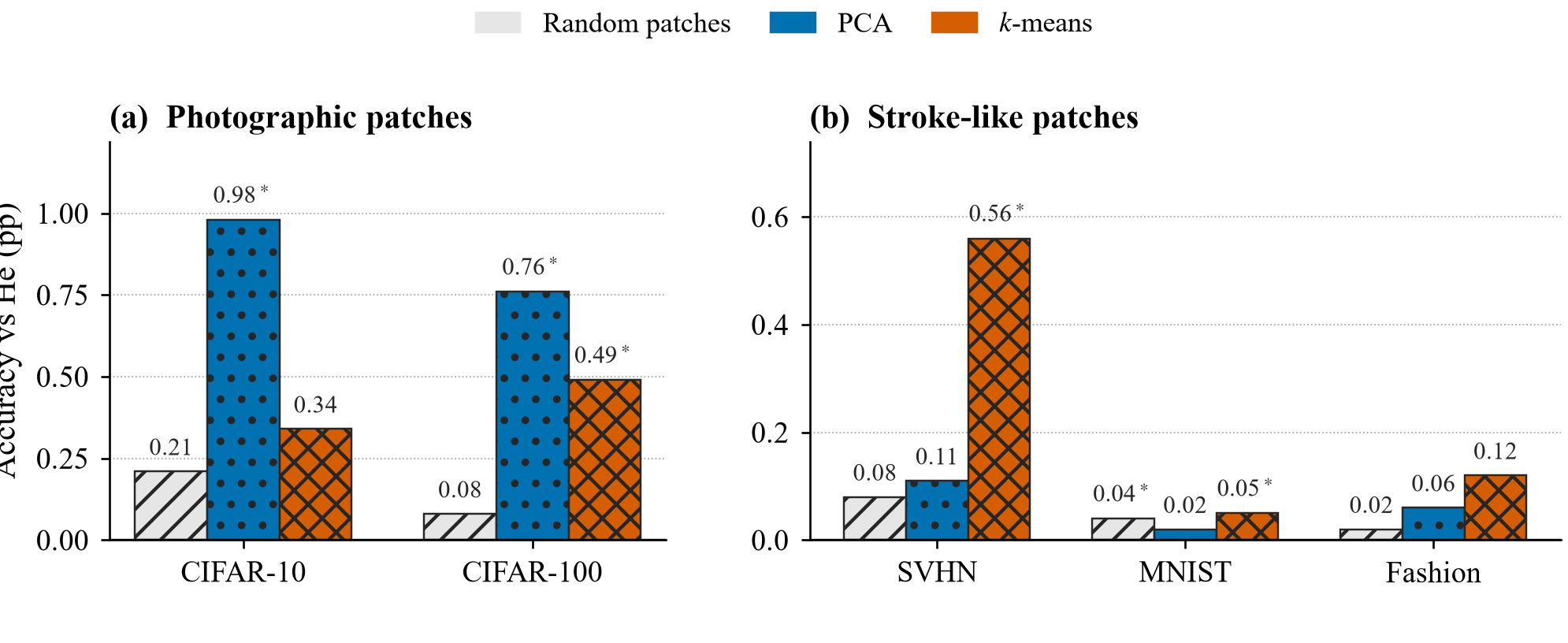}
\caption{Paired test-accuracy difference versus He initialization on FlexCNN ($n=8$
seeds). Asterisks mark $p<0.05$. The two panels use different $y$-scales. PCA is
the stronger dictionary on photographic patches; $k$-means is stronger on stroke-like
patches.}\label{fig:deltas}
\end{figure*}

The rule is only a first-batch check;
Figure~\ref{fig:deltas} and Table~\ref{tab:main} are the evidence.

\section{\uppercase{Limitations}}
\label{sec:limitations}

\noindent
Everything reported here is an initialization comparison on FlexCNN: two convolutions,
$5000$ Adam steps, and a write into the first half of conv1. That is enough to rank four
fills under a matched budget. It is not a claim about fully trained accuracy, and it is not
a claim about residual networks. On CIFAR-10 ResNet-20 \cite{he2016deep} the same
three dictionaries fell below He initialization. Batch normalization sits on the stem, so
a data-dependent fill is re-centered and re-scaled before it can matter, and conv1 is a
smaller piece of the model than it is in FlexCNN.

Table~\ref{tab:main} has the same qualification on the grayscale sets. Fashion-MNIST does not
reach $\alpha=0.05$ with eight seeds, and MNIST is already at $99.3\%$, so the
significant $k$-means increment there is not a useful gain. The patch count itself comes
from an occupancy coefficient fit to two training-set moments on these five datasets,
not from a foreground density measured independently of $n$. Pre-Warm only sets the
first-layer filters; it does not pretrain.

\section{\uppercase{Conclusions}}
\label{sec:conclusion}

\noindent
First-layer filters of a small convolutional network can be seeded from a single batch of
mean-centered patches by Pre-Warm: a rule-based fill that builds both a principal-component
dictionary and a $k$-means dictionary, sets the patch count from the filter count and a
foreground density, and writes the dictionary with the lower first-batch reconstruction
error into the first half of conv1. Under a matched FlexCNN protocol that split is
consistent across all five datasets. PCA is the stronger dictionary on the photographic
sets, CIFAR-10 and CIFAR-100. $k$-means is the stronger dictionary on the stroke-like
digit sets, SVHN and MNIST, and is the better of the two on Fashion-MNIST. Copying raw
patches does not reproduce the color-set gains. The two rules replace a validation search
over patch count and dictionary class, so the same first-batch procedure can be dropped
into an existing training run.

In future work we want that drop-in for larger models: a few lines of code, no change to
the architecture or the optimizer, and a modest but reliable boost rather than a new
training recipe. The fill remains both dictionaries, with the reconstruction check choosing
which one to write, mixed with standard random initialization as in Lehmann and Ebner
\cite{lehmann2021patches}.

\bibliographystyle{apalike}
{\small
\bibliography{refs}}

\end{document}